\pdfoutput=1

\documentclass[11pt]{article}

\usepackage[]{acl}

\usepackage{times}
\usepackage{latexsym}
\usepackage{array}
\usepackage{makecell}
\usepackage{booktabs}
\usepackage{multirow}
\usepackage{tabularx}
\usepackage[T1]{fontenc}

\usepackage[utf8]{inputenc}
\usepackage{indentfirst}
\usepackage{amsmath}
\usepackage{graphicx}
\graphicspath{{./image/}}
\usepackage{microtype}

\usepackage{inconsolata}

\title{MORE-3S:Multimodal-based Offline Reinforcement Learning \\ with Shared Semantic Spaces}

\author{Tianyu Zheng$^1$\;\quad Ge Zhang$^{2,3}$\thanks{Corresponding Author.}\;\quad Xingwei Qu$^{3}$,\; \\
    \textbf{Ming Kuang$^4$\quad Stephen W. Huang$^5$\quad Zhaofeng He$^{1}$\footnotemark[1]}\space
    \\[3pt]
    $^1$Beijing University of Posts and Telecommunications, $^2$University of Waterloo,\\ $^3$Multimodal Art Projection Research Community,$^4$ClouDr,$^5$Harmony.ai\\[3pt]
\vspace{-4ex}
\small
\texttt{zhengtianyu@bupt.edu.cn, ge.zhang@uwaterloo.ca} \\
}

\begin{document}

\maketitle

\begin{abstract}
Drawing upon the intuition that aligning different modalities to the same semantic embedding space would allow models to understand states and actions more easily, we propose a new perspective to the offline reinforcement learning (\textbf{RL}) challenge. 
More concretely, we transform it into a supervised learning task by integrating multimodal and pre-trained language models. 
Our approach incorporates state information derived from images and action-related data obtained from text, thereby bolstering RL training performance and promoting long-term strategic thinking.
We emphasize the contextual understanding of language and demonstrate how decision-making in RL can benefit from aligning states' and actions' representation with languages' representation.
Our method significantly outperforms current baselines as evidenced by evaluations conducted on Atari and OpenAI Gym environments. 
This contributes to advancing offline RL performance and efficiency while providing a novel perspective on offline RL.Our code and data are available at \url{https://github.com/Zheng0428/MORE_}.
\end{abstract}



\section{Introduction}
\noindent
RL operates as a sequential process where an agent perceives an environmental state, executes an action, observes the subsequent state, and receives a reward. 
Offline RL mitigates the limitations of conventional RL, like the need for long-term credit assignment bootstrapping and undesirable short-sighted learned policies, by learning policies from pre-existing datasets~\cite{Chen2021b}, eliminating the need for environmental interaction \cite{Levine2020a}.
This paradigm has attracted significant attention for its practical application potential in real-world RL training scenarios.

Unlike conventional offline RL techniques that primarily focus on learning value functions~\citep{Kostrikov2021} or policy gradients~\citep{Fujimoto2021}, recent approaches~\citep{Chen2021b,Schmied2023} view offline RL as a sequence modeling task. 
In these methods, past experiences of state-action-reward triplets are input into a transformer model~\citep{Vaswani2017}. 
They create a sequence of anticipated actions through a goal-conditioned policy, casting offline RL as a supervised learning problem. 
These approaches break away from the MDP assumption by using historical information to predict actions, effectively handling extensive sequences, and ensuring stability, which circumvents bootstrapping-related issues \cite{Kumar2020}.

However, it remains unclear how to effectively adapt Large-scale Pretrained Models (\textbf{LPM}) for offline RL \citet{Chen2021b}.
\citet{Reid2022a, Zhang2023a} demonstrate the effectiveness of fine-tuning LPM for offline RL tasks requiring high-level planning. 
Although further empirical evidence supports these findings \citep{Liu2022b, Fan2022}, the reasons behind these enhancements largely remain unexplored. 

To understand the underlying factors, we suggest two main considerations: first, the \textit{action rules}, such as grammatical structures or pragmatic rules essential for constructing valid sentences; second, the \textit{use cases}, implying the context where words and idioms are applied, especially when considering language as a complex game.

Within the RL and Natural Language Processing (\textbf{NLP}) contexts, we suggest that the latent representation of language may significantly contribute to RL decision-making processes. 
This interpretation views the next token prediction procedure in training Pre-trained Language Models (\textbf{PLMs}) as agents continuously making decisions, ascribing meaning to decisions through use. 
Therefore, integrating PLMs with RL may enhance an agent's ability to interpret and act meaningfully within its environment. 
Despite its theoretical appeal, empirical evidence supporting this hypothesis is sparse. 
A persistent challenge is decoupling high-level planning from low-level representations in the models \cite{Correia2022}. 
While preliminary attempts \citep{Lee2022b} show promising improvements based on decoupling, there is a pressing need for further research to unravel the mechanisms behind this phenomenon and to develop more effective methods to achieve it.

This study argues that the latent state representations derived from images during offline RL training, coupled with the discrete symbolic action space that models game operations, should correlate with their corresponding textual descriptions. 
\citet{Chen2021b, Reid2022a} directly utilize the latent state representations and discrete symbolic action space, without establishing alignment with their textual descriptions. 

Consider the game "Pong." In a specific frame, the left paddle is close to the top, the ball moves to the left, and the right paddle rises. This can be semantically described as: \textit{"Left paddle near top; ball moving left; right paddle rising"}. 
Traditional multimodal methods, which lack semantic grounding, might correlate raw pixels with actions like "move paddle up." 
However, the absence of a higher-level contextual understanding may reduce their resilience to minor game variations.
Grounding in a semantic space provides increased adaptability and learning efficiency.


Existing online RL methods~\citep{Wang2023c} usually seek to adapt LPM for high-level RL planning, introducing text-described states empirically without quantifying the advantage of modality alignment. 
This alignment allows the LPM to understand states and actions better. 
Nevertheless, the alignment of latent state representations, action space, and their textual descriptions facilitates agents initialized with LPMs' parameters to comprehend states and actions more effectively.
As a result, we propose a novel approach integrating multimodal and pre-trained sequence models into sequential RL, as shown in Fig.~\ref{fig:model}.
We design \textbf{M}ultimodal \textbf{O}ffline \textbf{R}einforcement l\textbf{E}earning with \textbf{S}hared \textbf{S}emantic \textbf{S}paces (\textbf{MORE-3S}), which demonstrates strong performance across various benchmark environments and offers improvements over existing offline RL methods. 
In summary, our primary contributions are three-fold:
\begin{figure*}[t]
\centering
\includegraphics[width=0.95\textwidth]{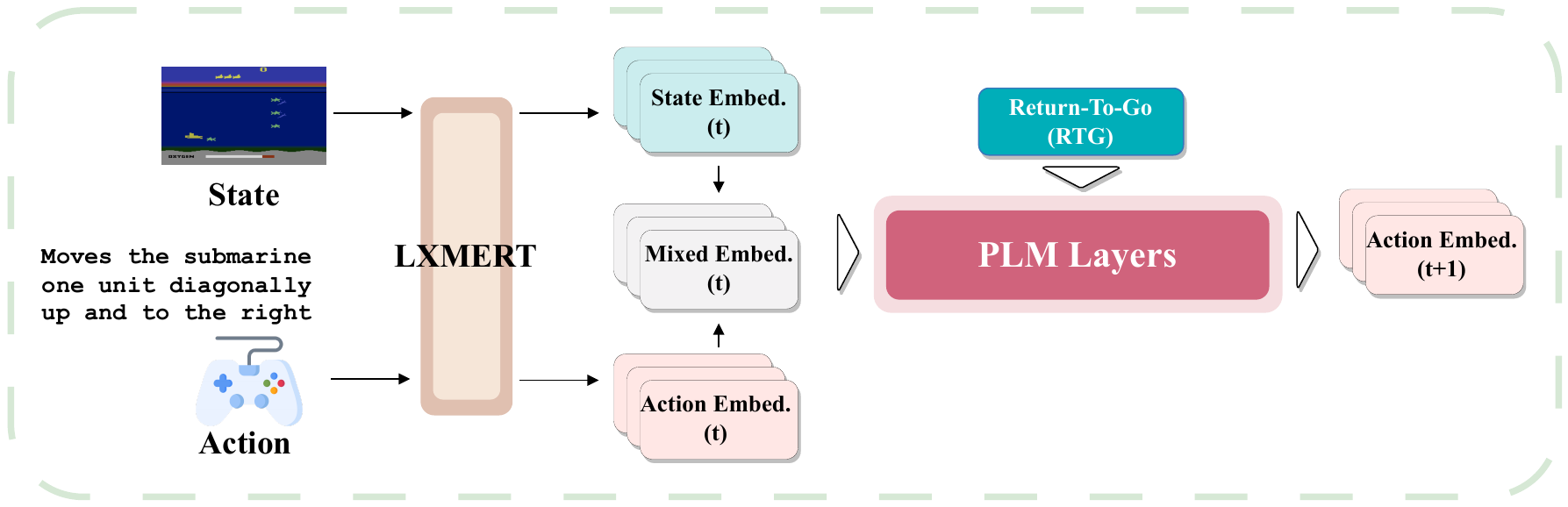}
\caption{Architecture diagram of the proposed MORE-3S approach. 
The Multimodal Encoder component combines the action (text) and state (image) inputs using the LXMERT model. ``Embed.'' denotes the embedding process. 
Autoregressive modeling of trajectories captures the system's dynamics by modeling trajectories as a sequence of tuples. 
LPMs predict subsequent actions based on the encoded sequence $O_t$, which corresponds to the 'Mixed Embed.' section in the diagram.}
\label{fig:model}
\end{figure*}
\begin{itemize}
\item We introduce a novel approach that aligns multimodal models with pre-trained sequence models for sequential RL, thereby enhancing RL training performance. 
\item MORE-3S, aligning states' and actions' representation with PLM's latent space, significantly improves the performance of offline RL algorithm. MORE-3S's performance is another convincing empirical observation proving the advances in high-level planning observed where NLP models are utilized in RL.
\item We propose integrating the \textit{return-to-go} into the attention mechanism of decision transformer-based RL models, enhancing their utilization of return-to-go.
\end{itemize}

\section{Related Work}
\noindent
\textbf{Offline Reinforcement Learning}\quad
Offline RL, a rapidly expanding field, focuses on deriving optimal policies from fixed datasets of trajectory rollouts, negating the need for further environmental interaction. 
These datasets, represented by $D={\left(s_{t}, a_{t}, r_{t}\right)}_{t=0}^{T}$, encompass states ($s_t$), actions ($a_t$), and rewards ($r_t$) for each timestep ($t$) until the episode time horizon ($T$), and are the products of a behavior policy and inherent dynamics. Offline RL, which utilizes value-based and model-based strategies, boasts a collection of noteworthy algorithms including BCQ \citep{Fujimoto2019}, AWR \citep{Peng2019}, BRAC \citep{Wu2019}, ICQ \citep{Yang2021}, CQL \citep{Kumar2020}, UWAC \citep{Wu2021}, MOPO \citep{Yu2020}, MOReL \citep{Kidambi2021}, Decision Transformer(DT)\citep{Chen2021b}, and Trajectory Transformer \citep{Janner2021}. 
Our approach innovatively integrates multimodal and pre-trained sequence models with sequential RL, aiming to overcome the limitations inherent in fixed-dataset policy optimization.

\noindent
\textbf{Sequence Modeling in RL}\quad
Research interest in the application of sequence modeling in RL has surged due to the use of Transformer-based decision models \citep{Hu2023a}. 
The key goal is to forecast future actions based on recent experiences. \citet{Chen2021b} applied a Transformer to a context-conditioned, model-free policy, while \citet{Janner2021} emphasized the sequence model's ability to predict states, actions, and rewards with beam search. 
Later studies fine-tune the Transformer for online environments \citep{Zheng2022}, and StAR-representations were developed for enhanced long-term modeling \citep{Shang2022a}.
Our proposed method improves decision transformer-based RL models by integrating return-to-go into the attention mechanism, enhancing data processing efficiency and offering a unique improvement over existing sequence modeling techniques.

\noindent
\textbf{Multimodal Data and Planning}\quad
Exploring the intersection of language with vision and text in RL is a recent trend, with studies increasingly incorporating NLP into the RL domain \citep{Sun2023,Liu2022c,Wang2023d,Jin24}.
This research area treats PLMs as offline RL agents and aims to separate high-level planning from modalities.

LPM for RL enables better reward descriptions, environment exploration, and agent communication, resulting in improved RL outcomes \citep{Huang2022a,Reed2022a,Ahn2022a,Liu2022b}.
Findings suggest a close link between language, local sequence relationships, and high-level planning across modalities \citep{Reid2022a, Tsimpoukelli2021, Zhang2022b}.
In this context, \citet{Liu2022b} explore a shared transformer architecture for vision and text games, bringing in the strength of NLP to the visual RL landscape.

Our work distinguishes itself in a crucial aspect: the implicit modality alignment. 
While previous studies have explicitly used pre-trained models like CLIP \citep{Radford2021,Lee2022b} for vision and language tasks in RL, our approach leverages implicit modality alignment by coupling latent state representations derived from images with their corresponding textual descriptions in the action space. 
This enables the model to achieve a more nuanced understanding of states and actions, ultimately enhancing RL performance. 
Unlike conventional approaches, which align modalities explicitly and in isolation \citep{jiang2020implicit,zeng2020aligntts}, our method inherently aligns these modalities through the training process. This novel implicit alignment allows for a more flexible integration of vision and text, addressing the complexity of real-world RL applications more effectively.

\section{Preliminary}
\noindent
\textbf{Trajectory Rollouts in Offline RL}\quad
In Offline RL, a trajectory is characterized as a sequence of state-action-reward triplets, represented as \(D = \{ (s_{t}, a_{t}, r_{t}) \}_{t=0}^{T}\). Here, \(s_t\) indicates the state, \(a_t\) the action executed, and \(r_t\) the reward received at time step \(t\). The dataset \(D\) comprises trajectories up to the time horizon \(T\), with each trajectory uniquely identified by index \(i\).

\vspace{0.1cm}
\noindent
\textbf{Sequence Representation in RL}\quad
In sequence modeling for RL, a sequence $\tau$ of experiences is used to predict the next actions. 
This sequence can be represented as follows:
\begin{align}
\tau = \left(r_{1}, s_{1}, a_{1}, r_{2}, s_{2}, a_{2}, \ldots, r_{N}, s_{N}, a_{N}\right)
\end{align}
This representation includes state-action-reward triplets and is suitable for offline RL and imitation learning scenarios, where the model can be trained in a supervised manner.
The goal of sequence modeling in RL is to predict prospective actions based on historical experiences. This can be formulated \citep{Janner2021}: 
\begin{align}
Pr(\hat{a_t}) = p(a_t| s_{1:t}, a_{1:t-1}, r_{1:t-1})
\end{align}

\noindent
\textbf{Transformers in RL}\quad
Transformers are neural network architectures that are highly successful in natural language processing tasks. 
Their ability to capture long-range dependencies in data makes them particularly suitable for sequence modeling in RL. 
In this context, Transformers can be used to model the relationships between state-action-reward triplets in trajectories and make predictions about future actions and rewards.

\vspace{0.1cm}
\noindent
\textbf{Multimodal Context in RL}\quad
Incorporating multimodal data such as text and vision can enhance the capabilities of RL agents. 
This can be particularly useful in complex environments where an agent needs to make decisions based on multiple sources of information. 
LPM can be adapted to function as Offline RL agents in a multimodal context, leveraging their capacity for high-level planning and local sequence modeling.


\section{Architecture and Training}
\noindent
In this section, we outline the MORE-3S approach. 
MORE-3S synergizes multimodal and pre-trained sequence models to enhance offline RL. 
Figure \ref{fig:model} illustrates the overall process, including state and action interpretation, as well as action prediction.

\subsection{Model Architecture}
\noindent
The model structure in this proposed Multimodal-based Offline Reinforcement Learning approach involves the integration of multimodal models and pre-trained sequence models into a Reinforcement Learning framework, where the multimodal model employs pre-training parameters to interpret state information from image data and action information from text, and the output of this multimodal model serves as the input for the LPM to predict subsequent actions.

\vspace{0.1cm}
\noindent
\textbf{Multimodal Encoder}\quad In the initial segment of MORE-3S, we delve into the two primary inputs at time $t$: the action and the state. 
These inputs are subsequently converted into their respective forms, with the action being transformed into text and the state into an image. 
To encode these inputs, we utilize the LXMERT \citep{Tan2019a} model, a robust multimodal transformer model pre-trained on a substantial corpus of image-text pairs. 
However, Nota Bene that MORE-3S's strength lies in the general approach of using a multimodal transformer model instead of the specific LXMERT. 
Upon encoding, the actions and states are amalgamated into a single sequence $O_t$, 
Specifically, 
\begin{align}
{O_t = \text{Concat}(E(A_t), E(S_t))}
\end{align}
where \( E(A_t) \) and \( E(S_t) \) represent the encoded forms of the action and the state at time \( t \), respectively.serving as the input for the subsequent component of our model. 

\vspace{0.1cm}
\noindent
\textbf{Autoregressive Modeling of Trajectories}\quad In the subsequent section of our study, we adopt the methodology delineated by \citet{Chen2021b}, wherein an autoregressive approach is employed to model trajectories. 
This necessitates the depiction of these trajectories as a sequence of tuples, formally represented as: 
\begin{align}
t=(\hat{R}_1, O_1, \hat{R}_2, O_2, . . . , \hat{R}_N , O_N )
\end{align}
Each tuple in this sequence encapsulates the returns-to-go ($\hat{R}_i$) and the decoupled information pertaining to the state and action ($O_i$) at a specific timestep $i$, given a total of $N$ timesteps.

This modeling strategy draws parallels with a sequence $x$, as explicated in the equation proposed by \citet{Bengio2000}: 
\begin{align}
P(\mathbf{x}) = \prod_{i  = 1}^{N} p\left(\mathbf{x}_{i} \mid \mathbf{x}_{i-1}, \mathbf{x}_{i-2}, \ldots, \mathbf{x}_{1}\right)
\end{align}
This equation embodies the fundamental principles of a self-regressive language model.
The returns-to-go, $\hat{R}_i$, is mathematically represented as $\hat{R}_{i}=\sum_{t=i}^{N} r_{t}$. 
By incorporating the processed information from the state and action ($O_N$) and the return-to-go ($R_N$) into our model, we effectively capture the dynamics of the system under investigation. 
This approach allows us to construct a comprehensive model that is capable of accurately representing the system's behavior over time.

\subsection{Memory Mechanism}
\noindent
To improve our model's proficiency in managing lengthy sequences, we integrated a memory mechanism inspired by \citet{Wu2022} into the GPT-style attention architecture. 
This memory approach enhances the model's ability to address long-term dependencies, as it facilitates storing and accessing information from prior stages of a sequence.

\begin{figure}[ht]
\centering
\includegraphics[width=0.48\textwidth]{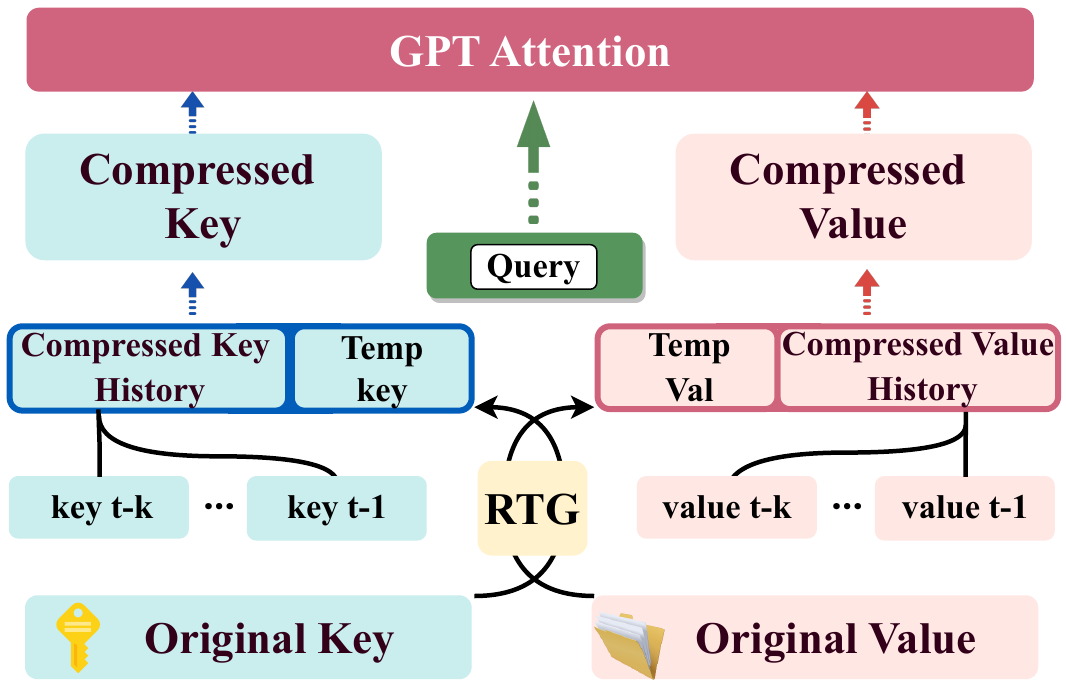}
\caption{Schematic Representation of the Integration of Return-to-Go (\textbf{RTG}) Quantity and Memory Mechanism in GPT-style Attention Architecture.}
\label{fig:attention}
\end{figure}

Incorporating the return-to-go quantity into the original keys $K$ and values $V$ of the transformer architecture has proven to be more effective than directly inputting it during the model's position embedding stage,see Figure \ref{fig:attention}.
This process leverages the attention weights at each time step $t$ to maintain a ``memory'' of earlier inputs, thus improving the model's capability to process tasks requiring long-term dependency handling.

The memory mechanism is practically implemented by extending the keys $\bar{K}$ and values $\bar{V}$ at the current step $t$ to include $\bar{K}(t')$ and $\bar{V}(t')$ from previous steps $t'$, yielding:
\begin{align}
\bar{V}^{t}= \left[\operatorname{sg}\left(\bar{V}^{(t-M)}\right), \cdot\cdot, \operatorname{sg}\left(\bar{V}^{(t-1)}\right), \bar{V}^{t}\right] \\
\bar{K}^{t}= \left[\operatorname{sg}\left(\bar{K}^{(t-M)}\right), \cdot\cdot, \operatorname{sg}\left(\bar{K}^{(t-1)}\right), \bar{K}^{t}\right] 
\end{align}
This formulation allows the query to attend information not only from the current time step $t$, but also from up to $M$ preceding steps, thereby offering an extended view(see Fiture \ref{fig:attention}). 
The increase in training and inference costs is primarily due to memory caching and the extended attention layer computation, but remains more efficient than traditional scaling methods, making our model practically suitable for diverse tasks.

\subsection{Training Details} 
\noindent
\textbf{Object-Level State Embeddings}\quad
Our approach involves utilizing position features and region-of-interest (\textbf{RoI}) features obtained from image-based states through a Fast R-CNN \citep{Girshick2015} network as input to our model. 
This approach differs from the conventional usage of feature maps generated by convolutional neural networks. 
Instead, we adopt the object feature-based embeddings proposed by \citet{Anderson2018}. 
By integrating object and position features, our method facilitates a more comprehensive comprehension of the image states, resulting in enhanced performance.

\vspace{0.1cm}
\noindent
\textbf{Semantic Embedding of Actions}\quad
To enhance the semantic comprehension of actions in the multimodal model LXMERT, we represent actions using both discrete and continuous textual descriptions. 
These representations facilitate a clearer understanding of actions, benefiting the model's comprehension and facilitating human interpretability.

As an exemplar, consider the actions within the game Breakout from the Atari 2600 suite:

\begin{itemize}
    \item \textbf{Action 0}: No action is taken, allowing the game to continue unchanged.
    \item \textbf{Action 1}: Launches a ball towards the bricks, aiming to break them.
    \item \textbf{Action 2}: Shifts the paddle to the right, intercepting the ball to prevent it from falling.
    \item \textbf{Action 3}: Shifts the paddle to the left, intercepting the ball to prevent it from falling.
\end{itemize}

For continuous environments based on the Mujoco framework, such as the HalfCheetah simulation, we provide textual prompts mapping specific dimensions of the action vector to torques or forces applied to particular joints. 
For instance, in the HalfCheetah environment, a prompt might state: "In the current time step, the Half-Cheetah’s six joints received the following torques or forces: First hind leg joint: a1, Second hind leg joint: a2, etc."

Each of these textual descriptions is generated from specific prompts, translating the numerical values in each dimension to the corresponding actions taken in the environment.

\subsection{Procedure of Training}
\noindent
In the training phase, the primary objective is to leverage the multimodal encoder in conjunction with the sequence model to accurately predict the subsequent timestep's output, \( O_{i+1} \). 
Importantly, the multimodal encoder is not subject to training; its parameters are fixed during this phase. 
The multimodal encoder processes the state and action data, represented as image and text respectively, at each timestep $i$, yielding an encoded output, $O_i$. This output, in tandem with the corresponding returns-to-go, denoted as $hat{R}_i$, serves as the input for the GPT-style sequence model.

Incorporating $\hat{R}_i$ into its attention mechanism, the sequence model can adjust its predictions contingent on the anticipated future returns. 
The model's primary task is to forecast the subsequent timestep's output, $O_{i+1}$, using the mean squared error (\textbf{MSE}) loss function as our objective. 
The MSE loss function is a natural choice for this supervised learning task, as it minimizes the squared differences between predicted and actual values. 
The predicted output distribution, $ O_{i+1}^{pred}$, is compared with the actual output distribution, $O_{i+1}$, to calculate the prediction error as follows:
\begin{align}
\mathcal{L} = \frac{1}{N}\sum_{i=1}^{N} (O_{i+1} - O_{i+1}^{pred})^2
\end{align}
By repeatedly minimizing the loss function $\mathcal{L}$, it becomes possible to incrementally enhance the model's predictions.

\section{Experiments}
\noindent
This section outlines our experimental setup, the benchmarks we employed, our evaluation metrics, and the implementation details of MORE-3S.
\subsection{Experimental Setup}
\noindent
Our experimental setup is constructed to evaluate our model in diverse environments, with a broad set of tasks that require different skill sets.
We employ Atari 2600 games~\citep{Bellemare2013} and OpenAI Gym environments~\citep{Brockman2016}, which encompass a wide range of complexity levels and task varieties, offering a comprehensive benchmark suite for our evaluation.

MORE-3S utilizes distinct pretraining models - LXMERT and a GPT-inspired model - for the optimization of RL training efficiency.
For LXMERT, the training spans an extensive range of datasets, hence fostering a comprehensive understanding of both visual and linguistic elements. 
We exclusively utilize the outputs of the Object Relationship Encoder and Language Encoder which are subsequently concatenated and integrated

For the GPT-inspired model, we use only the first six layers of the 12-layer GPT-2 model, allowing for a direct comparison with the DT and capitalizing on its strengths in sequence modeling for RL tasks \citep{Chen2021b}.
Pre-training of MORE-3S employs a model with 128 dimensions in its latent space, a solitary attention head, and six layers. 
The training, conducted with a learning rate of 3e-4 and a batch size comprising 65,536 tokens in each training batch, incorporates a warm-up schedule in the initial 10,000 steps.
We utilize the Byte-Pair Encoding \citep{Sennrich2016}, also used in GPT-2 \citep{Radford2019} and follow all hyperparameters given in \citep{Chen2021b}.


\subsection{Baselines}
\noindent
To augment our assessment of LPM, we conduct a comparative analysis against a suite of widely acclaimed offline RL algorithms. 
This collection of algorithms encompasses the DT, Conservative Q-Learning (CQL), Twin Delayed Deep Deterministic policy gradient enhanced with Behavior Cloning (TD3+BC), Behavior Regularized Actor Critic (BRAC), and Advantage Weighted Regression (AWR)  baselines.

\begin{table*}
\setlength{\tabcolsep}{4pt}
\centering
\small
\begin{tabular}{cccccccc}
\toprule
\textbf{Game} & \textbf{MORE-3S} & \textbf{CQL} & \textbf{QR-DQN} & \textbf{REM} & \textbf{BC} & \textbf{DT} & \textbf{StAR} \\
\midrule
\text{Breakout} & \textcolor{gray}{\textbf{300.5 $\pm$ 70.2}} & 211.1 & 17.1 & 8.9 & 138.9 $\pm$ 61.7 & 267.5 $\pm$ 97.5 & $\textbf{436.1}$ $\pm$ $\textbf{63.6}$ \\
\text{Qbert} & \textcolor{gray}{\textbf{60.8 $\pm$ 10.2}} & $\textbf{104.2}$ & 0 & 0 & 17.3 $\pm$ 14.7 & 15.4 $\pm$ 11.4 & 51.2 $\pm$ 11.5 \\
\text{Pong} & $\textbf{118.7} \pm \textbf{9.6}$ & \textcolor{gray}{\textbf{111.9}} & 18 & 0.5 & 85.2 $\pm$ 20.0 & 106.1 $\pm$ 8.1 & 110.8 $\pm$ 60.3 \\
\text{Seaquest} & $\textbf{3.1}$ $\pm$ $\textbf{0.3}$ & 1.7 & 0.4 & 0.7 & 2.1 $\pm$ 0.3 & \textcolor{gray}{\textbf{2.5 $\pm$ 0.4}} & 1.7 $\pm$ 0.3 \\
\bottomrule
\end{tabular}
\caption{\label{tab:atari}
Performance comparison of MORE-3S with models such as CQL, QR-DQN, REM, BC, DT, and StAR in four Atari games, using three seeds. 
MORE-3S's use of multimodal information and GPT-style pre-training led to high scores, highlighting its strong potential in gaming. Best mean scores are in bold and second-best mean scores are in gray bold.
}
\end{table*}

\begin{table*}[th]
\setlength{\tabcolsep}{4.5pt}
\centering
\small
\begin{tabular}{ccccccccccc}
\toprule
\textbf{Dataset} & \textbf{Environment} & \textbf{MORE-3S} & \textbf{CQL} & \textbf{BEAR} & \textbf{BRAC-v} & \textbf{AWR} & \textbf{MBOP} & \textbf{BC} & \textbf{DT} & \textbf{StAR} \\
\midrule
\multirow{3}{*}{\makecell{Medium\\Expert}} & HalfCheetah & 93.5 $\pm$ 1.0 & 62.4 & 53.4 & 41.9 & 52.7 & \textbf{105.9} & 59.9 & 86.8 $\pm$ 1.3 & \textcolor{gray}{\textbf{93.7 $\pm$ 0.1}} \\
 & Hopper & \textbf{112.2 $\pm$ 0.8} & 111.0 & 96.3 & 0.8 & 27.1 & 55.1 & 79.6 & 107.6 $\pm$ 1.8 & \textcolor{gray}{\textbf{111.1 $\pm$ 0.2}} \\
 & Walker & \textbf{111.3 $\pm$ 0.1} & 98.7 & 40.1 & 81.6 & 53.8 & 70.2 & 36.6 & 108.1 $\pm$ 0.2 & \textcolor{gray}{\textbf{109.0 $\pm$ 0.1}} \\
\midrule
\multirow{3}{*}{Medium} & HalfCheetah & 42.8 $\pm$ 0.1 & 44.4 & 41.7 & \textbf{46.3} & 37.4 & \textcolor{gray}{\textbf{44.6}} & 43.1 & 42.6 $\pm$ 0.1 & 42.9 $\pm$ 0.1 \\
 & Hopper & \textbf{76.4 $\pm$ 2.5} & 58.0 & 52.1 & 31.1 & 35.9 & 48.8 & 63.9 & \textcolor{gray}{\textbf{67.6 $\pm$ 1.0}} & 59.5 $\pm$ 4.2 \\
 & Walker & \textbf{81.5 $\pm$ 0.1} & 79.2 & 59.1 & \textcolor{gray}{\textbf{81.1}} & 17.4 & 41.0 & 77.3 & 74.0 $\pm$ 1.4 & 73.8 $\pm$ 3.5 \\
\midrule
\multirow{3}{*}{\makecell{Medium\\Replay}} & HalfCheetah & 36.5 $\pm$ 0.4 & \textcolor{gray}{\textbf{46.2}} & 38.6 & \textbf{47.7} & 40.3 & 42.3 & 4.3 & 36.6 $\pm$ 0.8 & 36.8 $\pm$ 3.3 \\
 & Hopper & \textbf{85.1 $\pm$ 2.7} & 48.6 & 33.7 & 0.6 & 28.4 & 12.4 & 27.6 & \textcolor{gray}{\textbf{82.7 $\pm$ 7.0}} & 29.2 $\pm$ 4.3 \\
 & Walker & \textbf{76.9 $\pm$ 1.5} & 26.7 & 19.2 & 0.9 & 15.5 & 9.7 & 36.9 & \textcolor{gray}{\textbf{66.6 $\pm$ 3.0}} & 39.8 $\pm$ 5.1 \\
\midrule
\textbf{Average} & & \textbf{79.6} & 63.9 & 48.2 & 36.9 & 34.3 & 47.8 & 46.4 & \textcolor{gray}{\textbf{74.7}} & 66.2 \\
\bottomrule
\end{tabular}
\caption{\label{tab:gym}
Results from the D4RL dataset reveal the average and variance across three seeds. 
Best mean scores are highlighted in bold. The second-best scores are highlighted in gray bold.
}
\end{table*}

\subsection{Atari}
\noindent
Our model is applied to the renowned Atari benchmark, a complex dataset developed by \citet{Bellemare2013}, to evaluate its performance in managing high-dimensional visual inputs and intricate credit assignments. 
Four games: Breakout, Pong, Qbert, and Seaquest from the Atari suite are used for the testing, following the evaluation design in \citep{Chen2021b,Agarwal2020}.

An offline dataset is constructed for the current study, consisting of approximately 1\% (or 500k transition steps) of the replay buffer dataset used in \citep{Agarwal2020}. 
Benchmark metrics are derived from various sources, including the Conservative Q-Learning (CQL), Random Ensemble Mixture (REM), and Quantile Regression DQN (QR-DQN) numbers from \citet{Kumar2020,Agarwal2020}. 
Additionally, we referenced imitation learning algorithms such as DT, StARformer, and BC.

Performance comparison of MORE-3S, with CQL, QR-DQN, REM, BC, DT, and StAR is detailed in Table \ref{tab:atari}. 
Remarkably, MORE-3S surpass all other models in Breakout and Qbert and achieved competitive results in Pong and Seaquest. 
This indicates the effectiveness of MORE-3S, which combines multimodal data during training and uses multimodal models to process visual and textual information. 
The model, initialized with pre-trained GPT-style parameters, shows significant improvement in Qbert and Breakout, emphasizing the value of incorporating multimodal information and pre-trained architectures in sequential reinforcement learning. 
This establishes the robustness of MORE-3S and underscores the successful implementation of pre-training and multimodal data utilization.

\subsection{OpenAI Gym}
\noindent
Our research employs the OpenAI Gym, a reputable platform that facilitates the development and comparison of RL algorithms across diverse environments. We evaluate three standard environments (HalfCheetah, Hopper, Walker), selected for their challenging control and decision-making characteristics. These are assessed using three dataset configurations, medium, medium-replay, and medium-expert, in line with the D4RL benchmark proposed by \citet{Fu2021a}. 
This benchmark standardizes scores for fair comparisons, where a score of 100 signifies an expert policy.


We evaluate our proposed model, MORE-3S, against several state-of-the-art algorithms to benchmark its performance. 
This includes model-free methodologies (CQL, BEAR, BRAC-v, AWR), a model-based approach (MBOP), and imitation learning algorithms (DT, StARformer). 
Performance metrics are derived from original or D4RL papers, and additional methods supplement our comprehensive evaluation. 
The experimental protocol comprise training models for 100k timesteps and conducting evaluations every 5k timesteps over 10 episodes. 
Using the formula \(100 \times (score - random~score) / (expert~score - random~score)\), normalized return scores are calculated to gauge MORE-3S's efficiency in continuous control tasks and its performance against leading algorithms.

Table \ref{tab:gym} illustrates MORE-3S's efficacy across D4RL's various environments. 
It demonstrates MORE-3S's superior performance on most tasks, with comparable results to the state-of-the-art algorithm on others. 
MORE-3S's unique feature is its integration of multimodal information processing mechanisms, which capacitates it to manage diverse inputs and structures. Pre-training a language model for reinforcement learning, reminiscent of GPT-style models \citep{Radford2019}, augments MORE-3S's capability. By leveraging the language models' knowledge via multimodal processing, MORE-3S effectively captures fine-grained spatial information and adapts to different environments. 
The results confirm MORE-3S's competitiveness and, in many cases, superiority to other state-of-the-art algorithms.


\section{Discussion}
\noindent
Through an exhaustive ablation study, we systematically dissect our computational model to understand the individual contributions of its various components. Our in-depth analysis evaluate the effects of implementing a condition-based return-to-go strategy, integrating pretrained GPT-style parameters, and applying LXMERT's pretrained parameters for state and action data preprocessing. 
We also examined the role of employing long-term attention mechanisms and assessed the implications of both context length and model size on overall performance.
Additionally, we conduct ablation studies across multiple games on the use of synonyms and contextual phrasing to explore the impact of varying action prompts. 
The experiment results indicate that varying action prompts can have little impact on MORE-3S's performance, which verifies MORE-3S's robustness.

To secure a robust evaluation, we execute a series of ablation experiments across computational environments. 
Our findings suggest that extending context length and model size beyond certain limits does not significantly enhance performance.

\vspace{0.1cm}
\noindent
\textbf{Condition-Based RTG}\quad
We examine two distinct strategies for integrating the return-to-go metric into our model: condition-based and linear layer integration. 
In the condition-based methodology, each temporal step $t$ processes the return-to-go through a linear layer, which is subsequently integrated into the transformer's native keys $K$ and values $V$. 
The linear layer approach, conversely, incorporates the return-to-go directly into the initial position embedding stage of the model.

The ablation study demonstrates a modest performance benefit of condition-based integration compared to the linear layer approach, as shown in Table \ref{tab:cbvslinear}. 
This modest advantage indicates that, while the condition-based method is superior in managing long-term dependencies, it does not produce markedly improved results. 
Consequently, the selection between the two methods should be based on the specific needs of the given RL task.

\begin{table}[ht]
\centering
\setlength{\tabcolsep}{4pt}
\resizebox{\linewidth}{!}{%
\begin{tabular}{lc}
\toprule
Model & Avg. Reward \\
\midrule
MORE-3S (Condition-Based) & 66.9 \\
MORE-3S (Linear layer integration)  & 65.4 \\
\bottomrule
\end{tabular}}
\caption{Comparison of condition-based and linear-layer integration of return-to-go.}
\label{tab:cbvslinear}
\end{table}

\vspace{0.1cm}
\noindent
\textbf{Randomizing Pretrained Parameters}\quad
\begin{figure}[b]
\centering
\includegraphics[width=0.48\textwidth]{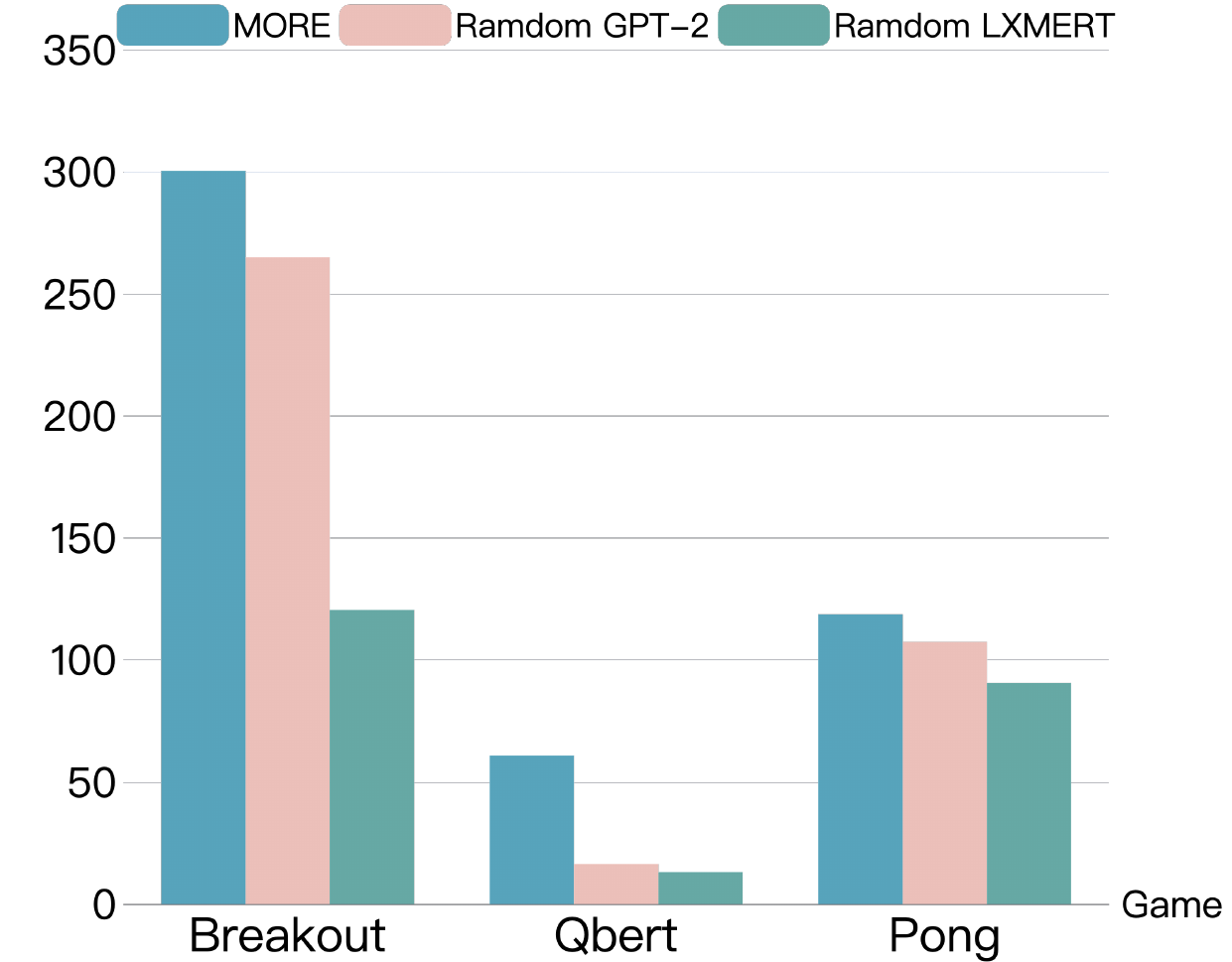}
\caption{Experiment on randomizing model weights versus finetuning them on OpenAI Gym.}
\label{fig:freeze_params}
\end{figure}
Our ablation study examines the effect of randomizing the pretrained parameters of GPT-style and LXMERT models on performance, as depicted in Figure \ref{fig:freeze_params}.
When the pretrained parameters of LXMERT are frozen, the model  demonstrates enhanced performance in processing multimodal data. 
This leads to improved interpretation of both state and action information, resulting in better performance across all evaluated environments compared to using randomly initialized LXMERT parameters.
Contrarily, when the GPT-style's pretrained parameters are absent, the model's initial performance is subpar, reminiscent of the DT\citep{Chen2021b} model. However, with sustained training, GPT-style's performance, although not superior, tends to converge towards that of our fully pretrained model.
This observation implies that even though pretraining GPT-style's parameters provides an initial performance boost, the model possesses the capability to effectively learn from scratch over time.
This finding further supports the robustness of our proposed method of integrating pretrained sequence models into sequential RL in diverse training conditions.

\vspace{0.1cm}
\noindent
\textbf{Long-Term Attention}\quad
We show the results of ablation studies comparing the performance of MORE-3S without long-term attention with full-fledged MORE-3S in Table \ref{tab:abl} to ascertain the efficacy of our proposed memory mechanism and the long-term attention module.

Our analysis illustrates that the model without a long-term mechanism often underperforms MORE-3S across numerous Atari games, affirming the memory mechanism and long-term attention's role in managing sequential data for RL tasks. 
Interestingly, MORE-3S without long-term components performs better on Seaquest, suggesting possible instances where long-term attention may not be vital. 
This observation necessitates further research. 


\begin{table}[ht]
\centering
\setlength{\tabcolsep}{4pt}
\resizebox{\linewidth}{!}{%
\begin{tabular}{lcc}
\toprule
Model & MORE-3S &MORE-3S\textbf{$\mathrm{_{ w/o \, Long-Term}}$ } \\
\midrule
Breakout & \textbf{300.5 ± 70.2} & 275.7 ± 66.9 \\
Qbert & \textbf{60.8 ± 10.2} & 55.3 ± 7.3 \\
Pong & \textbf{118.7 ± 9.6} & 105.2 ± 11.1 \\
Seaquest & 3.1 ± 0.3 & \textbf{3.3 ± 0.3} \\
\bottomrule
\end{tabular}}
\caption{The influence of the long-term attention component across different RL tasks.}
\label{tab:abl}
\end{table}

\vspace{0.1cm}
\noindent
\textbf{Varying Action Prompts in Game Environments}\quad
In our quest to discern the model's aptitude for linguistic comprehension, we venture into experiments with varied action prompt sets for each game, as showcased in Table \ref{tab:pm_combined_results}. 
Each game is subjected to three distinct action prompt categories:
\begin{itemize}
    \item \textbf{Original}: Action prompts as originally defined.
    \item \textbf{Synonyms}: Prompts incorporating synonymous replacements for pivotal action terms.
    \item \textbf{Contextual}: Prompts with deeper context or phrased as interrogative statements.
\end{itemize}

Taking the game Breakout as a reference, the prompt sets varied as:
\begin{itemize}
    \item Original: "Launches a ball towards the bricks, aiming to break them."
    \item Synonyms: "Fires a sphere at the blocks with the intent to shatter them."
    \item Contextual: "What would happen if a ball were propelled towards the bricks with the goal of breaking them?"
\end{itemize}

\begin{table}[htbp]
\centering
\small
\resizebox{\linewidth}{!}{%
\begin{tabular}{lcc}
\toprule
\textbf{Game} & \textbf{Action Prompt Set} & \textbf{Reward} \\
\midrule
\multirow{3}{*}{Breakout} & Original & 310.3 $\pm$ 68.1 \\
 & Synonyms & 295.4 $\pm$ 72.3 \\
 & Contextual & 305.2 $\pm$ 69.5 \\
\midrule
\multirow{3}{*}{Pong} & Original & 120.1 $\pm$ 8.5 \\
 & Synonyms & 117.9 $\pm$ 10.2 \\
 & Contextual & 119.5 $\pm$ 9.8 \\
\midrule
\multirow{3}{*}{Qbert} & Original & 58.5 $\pm$ 12.3 \\
 & Synonyms & 63.2 $\pm$ 11.1 \\
 & Contextual & 59.7 $\pm$ 10.5 \\
\midrule
\multirow{3}{*}{Seaquest} & Original & 3.0 $\pm$ 0.4 \\
 & Synonyms & 2.9 $\pm$ 0.5 \\
 & Contextual & 3.2 $\pm$ 0.2 \\
\bottomrule
\end{tabular}}
\caption{Results of Experiments with Different Action Prompt Sets}
\label{tab:pm_combined_results}
\end{table}

Upon closely examining the results, it is evident that there is minimal variance in performance across different prompt sets. Whether employing standard terminology, synonyms, or more context-rich queries, the model's performance is consistent. 
This uniformity underscores the model's linguistic proficiency and its adaptability to subtle linguistic variations in a gamified environment. 
These findings reinforce the reliability and robustness of our experimental method and suggest the model's superior linguistic interpretation in diverse scenarios.

\section{Conclusion}
\noindent
In conclusion, this research underscores the symbiotic relationship between RL and natural NLP, elucidating the tangible advantages conferred upon RL tasks when augmented with NLP models. 
Our pioneering MORE-3S method seamlessly amalgamates multimodal models with pre-trained sequence models in an RL milieu, delivering a compelling empirical corroboration for our theoretical postulations through demonstrably superior RL training outcomes. 
Instead of merely capitalizing on historical state-action-reward sequences,  our methodology elevates these dynamics by anchoring the actions in textually-described formats and the states in visual representations, thereby bolstering the model's precision and capability in interacting with its environment.
A noteworthy innovation we bring forth is the integration of the return-to-go parameter within the attention framework of decision transformer-centric RL models, further refining their performance. 
This scholarly endeavor illuminates potential avenues for an in-depth examination of the confluence between language models and reinforcement learning and heralds exciting prospects for the continued evolution of both offline RL and NLP domains.

\section*{Limitations}
Notwithstanding the promise demonstrated by our proposed method, it is crucial to recognize its constraints as they serve to inform future research.

\textbf{Model Interpretability}\quad While we have endeavored to integrate insights from Wittgenstein's language game theory, the opaque nature of our multimodal Transformer model may present interpretability issues. Consequently, tracing the reasoning behind the model's decisions can be challenging, which could potentially impede the application of our method in sectors where interpretability is indispensable.

\textbf{Limited Evaluation Environments}\quad The evaluation of our method was conducted in the Atari and OpenAI Gym environments. While these are common benchmarks in reinforcement learning (RL) research, they might not encapsulate the complexities and challenges inherent in more sophisticated environments, such as Minicraft. Consequently, the performance and applicability of our method in these multifaceted, multi-task environments remain largely untested, potentially limiting its generalizability.

\textbf{Model Complexity}\quad The introduction of pre-trained sequence models and multimodal models into the RL framework substantially amplifies the model's complexity. This escalation correlates with increased computational expenses and memory demands, which curtails the scalability of our method. Therefore, subsequent work should prioritize the development of more computationally economical strategies, whilst maintaining or even enhancing performance.

\textbf{Long-Term Planning}\quad Even though the memory mechanism enables our model to strategize for long-term decision-making, it remains uncertain how it performs in scenarios requiring elaborate long-term planning. It may prove inadequate for tasks characterized by extensive long-term dependencies.

Despite these constraints, we maintain that our research adds significant value to the field of reinforcement learning, by presenting an innovative multimodal RL approach inspired by language game theory. Future studies building on our work should address these constraints to further progress the effectiveness and applicability of RL systems.

\bibliography{custom}

\appendix
\section{Ablation}
\label{sec:ablation}
\subsection{Model Size}
Our exploration into the impact of altering the GPT-style model size, by varying parameters (117M, 345M, 774M), aimed to ascertain whether an increase in the model size would inevitably enhance its performance. Interestingly, our data, as presented in Table \ref{tab:modelsizecomparison}, suggested a possible overfitting occurrence with the escalating model size, thus challenging the idea of continual improvement with size augmentation. Notably, we observed diminishing relative gains as the parameter size escalated, potentially attributable to overfitting due to limited training data. This underscores a critical downside to overly emphasizing larger models, thereby stressing the need for a nuanced approach towards model size optimization. Nonetheless, this finding is stimulating, proposing that scaled language pre-training could yield benefits, particularly with larger and more diverse offline RL datasets in the future.

\begin{table}[ht]
\centering
\begin{tabular}{cccc}
\toprule
Model size & 117M & 345M & 774M \\
\midrule
Avg. Reward & 66.9 & 66.7 & 65.3 \\
\bottomrule
\end{tabular}
\caption{Comparison of different model sizes.}
\label{tab:modelsizecomparison}
\end{table}

\subsection{Trajectory Length}
Our comparative analysis of various trajectory lengths (1000, 500, and 100) during model training reveals no substantial enhancement in model performance with an increased trajectory length. This observation persists irrespective of the model's pretraining on long-range language modeling tasks, suggesting limited utility of elongated trajectories for tasks within the OpenAI Gym. There appears to be no significant advantage in terms of model robustness and efficiency derived from extended trajectories, thereby indicating a relative indifference of the model to the trajectory length, at least for the tasks under consideration.

\begin{table}[ht]
\centering
\begin{tabular}{cc}
\toprule
Context & Avg. Reward \\
\midrule
1000 & 66.9 \\
500 & 66.5 \\
100 & 67.3 \\
\bottomrule
\end{tabular}
\caption{Comparison of different contexts.}
\label{tab:contextcomparison}
\end{table}

\section{Action Prompt}
\label{sec:action}

This appendix provides comprehensive action prompts used in the training of the LXMERT multimodal model. Specifically, prompts were crafted for four distinct games: Breakout, Qbert, Pong, and Seaquest. These prompts function as accessible representations of discrete in-game actions, improving the model's understanding of action semantics. The forthcoming tables delineate action descriptions for each respective game.

\begin{table*}[ht]
\centering
\begin{tabular}{cc}
\toprule
\textbf{Action ID} & \textbf{Description} \\
\midrule
0 & No action is taken, allowing the game to continue unchanged. \\
1 & Launches a ball towards the bricks, aiming to break them. \\
2 & Shifts the paddle to the right, intercepting the ball to prevent it from falling. \\
3 & Shifts the paddle to the left, intercepting the ball to prevent it from falling. \\
\bottomrule
\end{tabular}
\caption{\label{tab:breakout}Breakout Action Descriptions}
\end{table*}

\begin{table*}[t]
\centering
\begin{tabular}{cc}
\toprule
\textbf{Action ID} & \textbf{Description} \\
\midrule
0 & Does nothing, keeping the game state unchanged. \\
1 & Launches the ball, initiating its movement. \\
2 & Moves the right paddle upward to block the ball and prevent it from crossing the right boundary. \\
3 & Moves the left paddle upward to block the ball and prevent it from crossing the left boundary. \\
4 & Launches the ball towards the right boundary, enabling the left player to score. \\
5 & Launches the ball towards the left boundary, enabling the right player to score. \\
\bottomrule
\end{tabular}
\caption{\label{tab:pong}Pong Action Descriptions}
\end{table*}

\begin{table*}[t]
\centering
\begin{tabular}{cc}
\toprule
\textbf{Action ID} & \textbf{Description} \\
\midrule
0 & Does nothing and allows the game to continue unchanged. \\
1 & Launches Qbert's ball to attack enemies. \\
2 & Moves Qbert one unit upward. \\
3 & Moves Qbert one unit to the right. \\
4 & Moves Qbert one unit to the left. \\
5 & Moves Qbert one unit downward. \\
\bottomrule
\end{tabular}
\caption{\label{tab:qbert}Qbert Action Descriptions}
\end{table*}

\begin{table*}[t]
\centering
\begin{tabular}{cc}
\toprule
\textbf{Action ID} & \textbf{Description} \\
\midrule
0 & Does nothing and maintains the game's current state. \\
1 & Launches a torpedo to attack enemies. \\
2 & Moves the submarine one unit upwards. \\
3 & Moves the submarine one unit to the right. \\
4 & Moves the submarine one unit to the left. \\
5 & Moves the submarine one unit downwards. \\
6 & Moves the submarine one unit diagonally up and to the right. \\
7 & Moves the submarine one unit diagonally up and to the left. \\
8 & Moves the submarine one unit diagonally down and to the right. \\
9 & Moves the submarine one unit diagonally down and to the left. \\
10 & Launches a torpedo upwards to attack enemies. \\
11 & Launches a torpedo to the right to attack enemies. \\
12 & Launches a torpedo to the left to attack enemies. \\
13 & Launches a torpedo downwards to attack enemies. \\
14 & Launches a torpedo diagonally up and to the right to attack enemies. \\
15 & Launches a torpedo diagonally up and to the left to attack enemies. \\
16 & Launches a torpedo diagonally down and to the right to attack enemies. \\
17 & Launches a torpedo diagonally down and to the left to attack enemies. \\
\bottomrule
\end{tabular}
\caption{\label{tab:seaquest}Seaquest Action Descriptions}
\end{table*}

\end{document}